\documentclass[letterpaper]{article}
\usepackage[top=0.85in,left=2.75in,footskip=0.75in,marginparwidth=2in]{geometry}

\usepackage[utf8]{inputenc}

\usepackage{cite}

\usepackage{nameref,hyperref}

\usepackage[right]{lineno}

\usepackage{microtype}
\DisableLigatures[f]{encoding = *, family = * }

\raggedright
\usepackage{changepage}

\usepackage[aboveskip=1pt,labelfont=bf,labelsep=period,singlelinecheck=off]{caption}

\makeatletter
\renewcommand{\@biblabel}[1]{\quad#1.}
\makeatother

\usepackage{lastpage,fancyhdr,graphicx}
\usepackage{epstopdf}
\fancyheadoffset[L]{2.25in}
\fancyfootoffset[L]{2.25in}

\usepackage{color}

\definecolor{Gray}{gray}{.25}

\usepackage{graphicx}

\usepackage{sidecap}
\usepackage{url}
\usepackage{wrapfig}
\usepackage[pscoord]{eso-pic}
\usepackage[fulladjust]{marginnote}
\usepackage{amsmath}
\reversemarginpar

\begin{document}
	\vspace*{0.35in}
	
	% title goes here:
	\begin{flushleft}
		{\Large
			\textbf\newline{Is Neuromorphic MNIST neuromorphic? Analyzing discriminative power of neuromorphic datasets in the time domain }
		}
		\newline
		% authors go here:
		\\
		Laxmi R Iyer\textsuperscript{1},
		Yansong Chua\textsuperscript{1,*}
		Haizhou Li\textsuperscript{2},
		\\
		\bigskip
		\bf{1} Institute of Infocomms Research, A*Star, Singapore
		\\
		\bf{2} Department of Electrical and Computer Engineering, National University of Singapore, Singapore
		\\
		\bigskip
		* chuays@i2r.a-star.edu.sg
		
	\end{flushleft}
	
	\section*{Abstract}
	
	The advantage of spiking neural networks (SNNs) over their predecessors is their ability to spike, enabling them to use spike timing for coding and efficient computing. A neuromorphic dataset should allow a neuromorphic algorithm to clearly show that a SNN is able to perform better on the dataset than an ANN. We have analyzed both N-MNIST and N-Caltech101 along these lines, but focus our study on N-MNIST. First we evaluate if additional information is encoded in the time domain in a neuromoprhic dataset. We show that an ANN trained with backpropagation on frame based versions of N-MNIST and N-Caltech101 images achieve 99.23\% and 78.01\% accuracy. These are the best classification accuracies obtained on these datasets to date. Second we present the first unsupervised SNN to be trained on N-MNIST and demonstrate results of 91.78\%. We also use this SNN for further experiments on N-MNIST to show that rate based SNNs perform better, and precise spike timings are not important in N-MNIST. N-MNIST does not, therefore, highlight the unique ability of SNNs. The conclusion of this study opens an important question in neuromorphic engineering – what, then, constitutes a good neuromorphic dataset?  
	% now start line numbers
	%\linenumbers

% the * after section prevents numbering
\section{Introduction}

The remarkable performance and efficiency of the brain have prompted scientists to build systems that mimic it. Early neural networks, networks of the first and second generations do not have neurons that spike. These networks, known as artificial neural networks (ANNs) have real number outputs that are between 0 and 1 and can be seen as time averaged firing rates of neurons. The networks of the third generation, known as spiking neural networks (SNN) explicitly employ spikes as their mechanism for computation. Third generation networks are more mathematically accurate models of biological neurons. A neuron of the third generation network receives incoming current through its synapses and fires a spike when its membrane potential exceeds a threshold. Such a neuron can use spike time coding, described below. Before we describe spike time coding, we will first enumerate the different definitions of firing rate currently used. 

\begin{figure}[ht] %s state preferences regarding figure placement here
	
	% use to correct figure counter if necessary
	%\renewcommand{\thefigure}{2}
	
	\includegraphics[width=\textwidth]{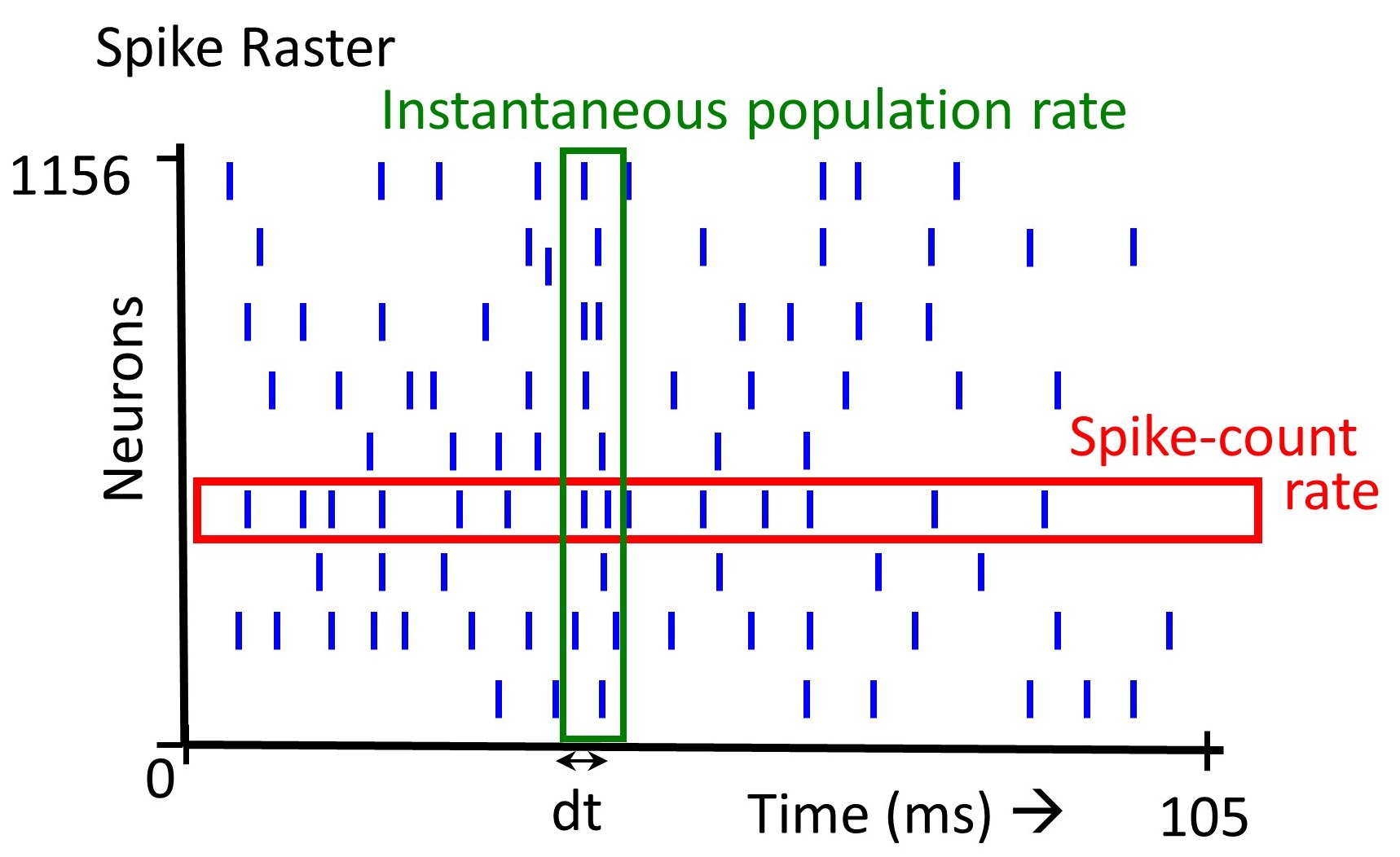}
	
	\caption{\color{Gray} \textbf{Spike-rate definition.}. There are several definitions of spike rate. Averaging over all the spikes emitted by a single neuron (spikes in the red box), we get the \emph{spike-count rate}. Averaging over the spikes emitted at a time instant by all the neurons (spikes in the green box), we get the \emph{instantaneous population rate.}}
	
	\label{fig:rate} % \label works only AFTER \caption within figure environment
	
\end{figure}

The firing rate of a spiking neuron is defined in several ways  – 1. The time averaged firing rate is the number of spikes fired by a neuron over a certain duration,  2. The instantaneous population firing rate is the number of spikes elicited by a population of neurons in a small time window, 3. The trial averaged firing rate of a neuron firing is the average number of spikes across trials. Note that definition 2 and 3 denote firing rate as a variable in time. The first two definitions are illustrated in Figure \ref{fig:rate}. In this paper, we focus primarily on the first definition (Figure \ref{fig:rate}A), but also consider the second definition (Figure \ref{fig:rate}B). 

Scientists have debated over how neurons code information - whether the brain follows a rate code or a temporal code (\cite{Brette:2015}). Rate code makes use of the firing rate of neurons while temporal code makes use of the precise spike timing of the neurons.  The issue of time and rate coding, as summarized in \cite{Brette:2015} is as follows: does spike firing rate of a neuron capture most of the important information and computations, rendering the exact timing of spikes unnecessary? 

Several studies have highlighted the importance of precise spike times. Firstly, \cite{Gerstner:1999} shows that there are specialized subsystems for which the precise timing of spikes are relevant.  The specialized subsystems include the electrosensory system of electric fish (\cite{Heiligenberg:1991}, \cite{Metzen:2016}) and the auditory system of barn owls (\cite{Gerstner:1999}, \cite{Carr:1990}, \cite{Konishi:1993}, \cite{Wagner:2005}, \cite{Keller:2015}, \cite{Carr:2016}). Behavioral experiments on owls show that they can locate sound sources in complete darkness with extreme precision. They can detect a temporal difference of around 5$\mu$$s$ between the left and right ear. Such precise calculations invalidate the use of an averaging mechanism in the brain. Secondly, \cite{Thorpe:2001} details several arguments for spike time codes. Experiments show that primates are able to perform visual classification as fast as 100-150$ms$ after the stimulus is presented (\cite{Thorpe:2001}, \cite{Kirchner:2006}, \cite{Butts:2007}, \cite{Crouzet:2010}). Given that this information must have passed about 10 layers of processing, each individual processing stage is completed on average in only 10 $ms$, rendering a rate coding mechanism highly unlikely (\cite{Thorpe:2001}, \cite{Butts:2007}). Further, the number of photoreceptors that are present in the retina and the resolution of the images processed invalidate an instantaneous population rate code \cite{Thorpe:2001}. Thorpe examines both spike-count rate and instantaneous population rate coding using Poisson spikes, the most prevalent rate coding scheme. Through simple statistical analysis he demonstrates that Poisson coding is not efficient enough to transmit detailed information about the level of excitation in a sensory receptor - and there are several studies detailing the importance precise spike times in sensory systems - 1.) \cite{Johansson:2004} points out that precise timing of the first spikes in tactile afferents encodes touch signals. Tactile perception is shaped by millisecond precise spike timing (\cite{Mackevicius:2012}, \cite{Saal:2015}). 2.) In cats and toads, retinal ganglion cells encode information about light stimuli by firing only 2-3 spikes in 100 ms (\cite{Gabbiani:2001}). 3.) Studies have also shown the importance of spike timing in the vestibular system (\cite{Sadeghi:2007}) and somatosensory cortex (\cite{Harvey:2013}, \cite{Zuo:2015}). Finally, results in neuroprosthetics show that precise relative timing of spikes is important in generating smooth movement (\cite{Popovic:2000}).  These studies suggest that when high speed of a neural system is required, timing of individual spikes is important. With the importance of precise spike timings, there are several neural coding theories that take spike timing into account - examples are time to first spike (\cite{Johansson:2004}, \cite{Saal:2009}), rank order coding (\cite{Thorpe:2001}, \cite{VanRullen:2001}, \cite{Kheradpisheh:2017}), polychronization (\cite{Izhikevich:2006}), coding by synchrony (\cite{Gray:1989}, \cite{Malsburg:1999}, \cite{Singer:1999}), predictive spike coding (\cite{Deneve:2008}). 

Spike time coding does not need a large number of spikes or many neurons to quantify large values, but can do so by varying the spike timing of a few neurons. As a result, spike time codes allow more efficient computation. If a SNN is just using time-averaged or instantaneous population rate codes, it would be less efficient than ANNs, as it would need to run for long periods of time or employ many neurons to compute accurate averages of spike rates. The main advantage of the SNNs over the previous two generations of neural networks is that they can, in principle, employ spike time coding for higher efficiency. 

Along with the advances of neuromorphic engineering, there arises the need for a neuromorphic dataset to benchmark different SNNs. In Computer Vision, MNIST and Caltech101 are examples of well known image datasets. Recently neuromorphic algorithms have been tested against MNIST (e.g.\cite{Querlioz:2013}, \cite{Diehl:2015}, \cite{Kheradpisheh:2017}). To do this, images are converted to spikes using different methods. For e.g., \cite{Querlioz:2013} and \cite{Diehl:2015} convert images to Poisson spike trains with spike rates proportional to the intensity of the pixels. \cite{Kheradpisheh:2017} convert images to spikes with spike times proportional to image contrast. However, to advance the field of neuromorphic algorithms,  a dataset whereby features are encoded in asynchronously in time is required, which incidentally renders any data pre-processing unnecessary. N-MNIST, N-Caltech101 \cite{Orchard:2015}, MNIST-DVS and CIFAR10-DVS  are datasets recorded by moving either a vision sensor or the image from a pre-existing Computer Vision dataset and recording the resultant images. For example, Neuromorphic MNIST (N-MNIST) and Neuromorphic Caltech101 (N-Caltech101) \cite{Orchard:2015} are recorded by moving an ATIS vision sensor (\cite{Posch:2011}) across the original MNIST and Caltech101 patterns respectively in 3 predefined directions. The ATIS vision sensor is a neuromorphic sensor that records pixel-level intensity changes in the scene, based on the principles of the retina. The N-MNIST and N-Caltech101 patterns are therefore, represented as events occurring at pixel locations. The N-MNIST dataset has been successfully tested on many recent neuromorphic algorithms (for e.g. \cite{Cohen:2016}, \cite{Lee:2016}, \cite{Neil:2016a}, \cite{Neil:2016b}). However, to test the ability of third generation spiking neural networks, ideally, a neuromorphic dataset used should contain discriminatory features in the time domain. We want to examine if these neuromorphic datasets satisfy this criterion by integrating information over time as described below. 

\cite{Orchard:2015} mentions that in N-MNIST and N-Caltech101, the movement of the ATIS sensor mimics retinal saccades. However, our visual system is designed to extract information about the 3D world from many 2D image projections formed by the retina (\cite{Elder:2016}). Visual information is integrated across retinal saccades (\cite{Fiser:2002}) to provide a more holistic visual representation, for example to group visual input to separate image from ground (\cite{Blake:2005}). In addition, as \cite{George:2008} describes, we are very adept at recognizing images despite different rotations, scales, and lighting conditions (also \cite{Simoncelli:2003}). Such an integrated representation of objects is obtained from data varying continuously in time over all these different dimensions, in ways that conform to laws of physics (\cite{Blake:2005}, \cite{George:2008}, \cite{Mazzoni:2011}, \cite{Lake:2016}, \cite{Keitel:2017}). Therefore, time is probably acting as a supervisor providing useful information to enable us to create such a holistic representation \cite{George:2008}. It is therefore necessary to ask if saccadic movements of the camera used to record N-MNIST and N-Caltech101 gather information that is just as rich and critical for classfication. Saccades in these datasets are constructed by moving a camera over 2D static images in a predefined manner. This may not match the description of retinal saccades given by \cite{Fiser:2002}, \cite{George:2008}. At the very least it should provide additional information from the original MNIST and Caltech101. We therefore want to know what role time plays in these datasets.
We commence our study with both N-MNIST and N-Caltech101, but focus the rest of this study on N-MNIST alone. 
% several questions here. But yes, perhaps can look for citations. Ok other citations. 
Therefore, in this paper, we ask two questions about neuromorphic datasets recorded from pre-existing Computer Vision datasets by moving the images or a vision sensor: 

\begin{enumerate}
	\item Does the timing of spikes in these neuromorphic datasets provide any useful information? 
	\item Do these neuromorphic datasets highlight the strength of SNNs? 
\end{enumerate}

These questions are important from various viewpoints – from a general machine learning perspective, we want to know if these neuromorphic datasets can be classified by ANNs  just as well, or even more efficiently. From the neuromorphic perspective, a neuromorphic dataset should be able to highlight the unique properties and strengths of SNNs over ANNs in certain machine learning tasks. From the neuroscience point of view, it would be interesting to investigate if this method of recording from static images would gather additional information in the time domain than that available in the original Computer Vision datasets (such as MNIST and Caltech101), which can then be further utilized by some learning algorithms.  

To address these questions, we present several experiments with the neuromorphic datasets. We start off with a description of N-MNIST and N-Caltech101 datasets after which we describe our first experiment. Here, N-MNIST and N-Caltech101 are trained on an ANN. We then describe a design space that further experiments would explore, followed by other experiments that compare the performance of temporal and rate based SNNs on the N-MNIST dataset. This is followed by an experiment that classifies the N-MNIST dataset using an SNN trained with a data-derived STDP rule based on instantaneous population rates. Finally we conclude with a discussion on the implications of these results, and other related questions. 

\section{N-MNIST and N-Caltech101 Data Format}

The N-MNIST dataset is created by moving the ATIS vision sensor over each  MNIST image. This is done for all 60000 training images and 10000 test images in MNIST. The camera has 3 pre-defined movements (or \emph{saccades}).  Each N-MNIST spike train is $360$ ms long – divided into 3 saccades. The first saccade occurs during the first $105$ ms (0-105 ms), the second saccade in the next $105$ ms (105-210 ms), and the third saccade in the next $105$ ms (210-315ms)\footnote{\url{https://github.com/gorchard/Matlab_AER_vision_functions/}} (\cite{Cohen:2016}). Finally there is a $45$ ms additional time appended to end of 315ms to ensure that the last events have an effect on learning (\cite{Cohen:2016}).

N-MNIST patterns are represented as \emph{events}, each occurring at a specific pixel location or \emph{address} at a particular time (each event has a time stamp in $\mu$s). This is known as the \emph{address-event representation (AER)} protocol. Events elicited due to an increase in pixel intensity are characterized as $ON$ events, and decrease in pixel intensity, as $OFF$ events. 

In our experiments we consider $ON$ events in the first saccade (0-105 ms) only. We reduce the time resolution of the spike trains by binning events with $\mu$s time stamp into ms intervals. In this paper, each pattern refers to the events recorded during a single saccade. 

Caltech101 contains $8709$ images, and N-Caltech101 is created in the same manner from the ATIS vision sensors.  

\section{Experiment: Training N-MNIST and N-Caltech101 images with an Artificial Neural Network} \label{sec:ANN_TF}

This experiment examines the performance of \emph{time-collapsed} versions of N-MNIST and N-Caltech101 on artificial neural networks (ANN). 

\begin{figure}[ht] %s state preferences regarding figure placement here
	
	% use to correct figure counter if necessary
	%\renewcommand{\thefigure}{2}
	
	\includegraphics[width=\textwidth]{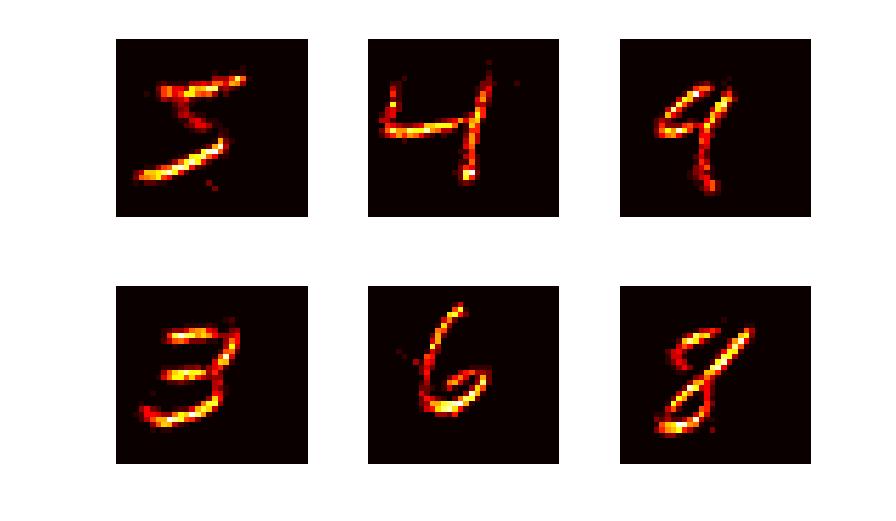}
	
	\caption{\color{Gray} \textbf{N-MNIST time collapsed images.}. N-MNIST patterns are collapsed in the time dimension to static images with pixel intensity proportional to the event rate of the pixel. These images are trained on an ANN to examine the removal of the temporal component in N-MNIST affects the performance. The above are 6 such images created from N-MNIST time-collapsed patterns.}
	
	\label{fig:sqImg} % \label works only AFTER \caption within figure environment
	
\end{figure}

In this experiment N-MNIST and N-Caltech101 patterns are collapsed in the time dimension to static images with pixel intensity proportional to the spike rate of the pixel \ref{fig:sqImg}. The conversion from AER to static images is done as follows. Each pattern $p$ can be represented as a set of spike trains, one for each pixel. The spike train for pattern $p$, pixel $x$ is $s^{x, p} = \{t^{x, p}_1, t^{x, p}_2, ... t^{x, p}_n \}$ where each element denotes the time of spike. Note that $t^{x, p}_1, ... , t^{x, p}_n$ are in the range $[0, 105]ms$ since we consider only saccade $1$ ($ON$ polarity). The normalized spike counts $C^{x, p}$ are calculated as follows: 

\begin{equation}
C^{x, p} = \frac{\sum_i^n t^{x, p}_i}{\max_y \sum_i^n t^{y, p}_i}
\end{equation}  

So $C^{x, p}$ is normalized by the highest spike count per pixel in pattern $p$. Note that spike counts are normalized per pattern, so patterns with low spike rates have their overall $C^p$, i.e. normalized spike count vector for a pattern, increased.

Each \emph{time collapsed} N-MNIST image pattern $p$ is a $34 \times 34$ image with intensity values at each pixel $x$ being $C^{x, p}$. The patterns are trained in Keras on a convolutional neural network (CNN). The first two convolutional layers both consist of $32$ neuronal maps of size $3 \times 3$ each with ReLU activation. The first convolutional layer has same padding. The convolutional layers are followed by a max pooling layer with a size of $2 \times 2$, after which a Dropout with rate $0.25$ is applied. This is followed by two convolutional layers both consisting of $64$ neuronal maps of size $3 \times 3$ each with ReLU activation. This is followed by a max-pooling layer that has a size of $2 \times 2$ after which a dropout of rate $0.25$ is applied. 

This is followed by a fully connected layer of $128$ output neurons, after which a dropout with rate $0.5$ is applied. This is followed by another final fully connected layer with $10$ output neurons and softmax activation which does the classification. The loss function used is cross entropy, and the Adadelta optimizer is applied. After running 100 epochs, we get an accuracy of $99.23\%$. We compare this to the performance of other state-of-the-art algorithms on N-MNIST in the table below.  %write the table number. 

\begin{table}[!ht]
	%\begin{adjustwidth}{-1.5in}{0in} % comment out/remove adjustwidth environment if table fits in text column.
	\centering
	\caption{{\bf Comparison of N-MNIST performance.} In this table a comparison of our method on N-MNIST with other state-of-the-art algorithms is given below. }
	\begin{tabular}{|c|c|}
		\hline
		\hline
		Method & N-MNIST  \\
		\hline
		SNN on Backprop (\cite{Lee:2016}) & 98.66\% \\
		\hline
		HATS (\cite{Sironi:2018}) & 99.1\% \\
		\hline
		Active Perception with DVS )(\cite{Yousefzadeh:2018}) & 98.8\% \\
		\hline
		\textbf{Collapsed images with ANN} & \textbf{99.23\%} \\
		\hline
		%\multicolumn{4}{|l|}{\bf Heading 1} & \multicolumn{3}{|l|}{\bf Heading 2}\\ \hline
		%cell 1 - row 1 & cell 2 - row 1 & cell 3 - row 1 & cell 4 - row 1 & cell 5 - row 1 & cell 6 - row 1 & cell 7 - row 1 \\ \hline
		%cell 1 - row 2 & cell 2 - row 2 & cell 3 - row 2 & cell 4 - row 2 & cell 5 - row 2 & cell 6 - row 2 & cell 7 - row 2 \\ \hline
		%cell 1 - row 3 & cell 2 - row 3 & cell 3 - row 3 & cell 4 - row 3 & cell 5 - row 3 & cell 6 - row 3 & cell 7 - row 3 \\ \hline
	\end{tabular}
	\label{tab1}
	%\end{adjustwidth}
\end{table}

N-Caltech101 has images of different sizes. Each \emph{time collapsed} N-Caltech101 image pattern $p$ is resized to a $224 \times 224$ image. These images are trained on a VGG-16 convolutional neural network pretrained on ImageNet. The methodology used for training is detailed in another paper by our group \cite{Gopalakrishnan:2018}, where we examine N-Caltech101 more thoroughly. A comparison of N-MNIST and N-Caltech101 performance on several algorithms is given in the table below. 

\begin{table}[!ht]
	%	\begin{adjustwidth}{-1.5in}{0in} % comment out/remove adjustwidth environment if table fits in text column.
	\centering
	\caption{{\bf Comparison of N-MNIST and N-Caltech101} In this table a comparison of our methods and other well-known algorithms on N-MNIST and N-Caltech101 are given. }
	\begin{tabular}{|c|c|c|}
		\hline
		Method & N-MNIST & N-Caltech101 \\
		\hline
		H-First (\cite{Orchard:2015b}) & 71.2 & 5.4 \\
		\hline
		HOTS (\cite{Lagorce:2017}) & 80.8 & 21.0 \\
		\hline 
		Gabor-SNN (\cite{Sironi:2018}) & 83.7 & 19.6 \\
		\hline
		HATS (\cite{Sironi:2018}) & 99.1 & 64.2 \\
		\hline
		\textbf{Collapsed images with ANN} & \textbf{99.23\%} & \textbf{78.01 \%}\\
		\hline
	\end{tabular}
	\label{tab1}
	%	\end{adjustwidth}
\end{table}

Before this paper, HATS algorithm \cite{Sironi:2018} gave the best performance accuracy on event-based datasets. However, the collapsed images on an ANN perform better than the HATS method. Our method of just summing up spikes over time (therefore getting rid of the time representation) is able to obtain the best accuracy ever obtained on these neuromorphic datasets. In N-Caltech101, it has a 10\% increase in accuracy over the previous methods. This in turn indicates that time does not provide any useful information to assist in classification. 

\section{Spiking Neural Network}

The rest of the experiments in this paper are run on spiking neural networks (SNN).  In this section we will describe the SNN that is used for the experiments. The SNN algorithm in this paper closely follows \cite{Diehl:2015}, but has been modified to suit the N-MNIST dataset. For a detailed description of these modifications, refer to \cite{Iyer:2017}. 

\subsection{Network Architecture}

\begin{figure}[ht] %s state preferences regarding figure placement here
	
	% use to correct figure counter if necessary
	%\renewcommand{\thefigure}{2}
	
	\includegraphics[width=\textwidth]{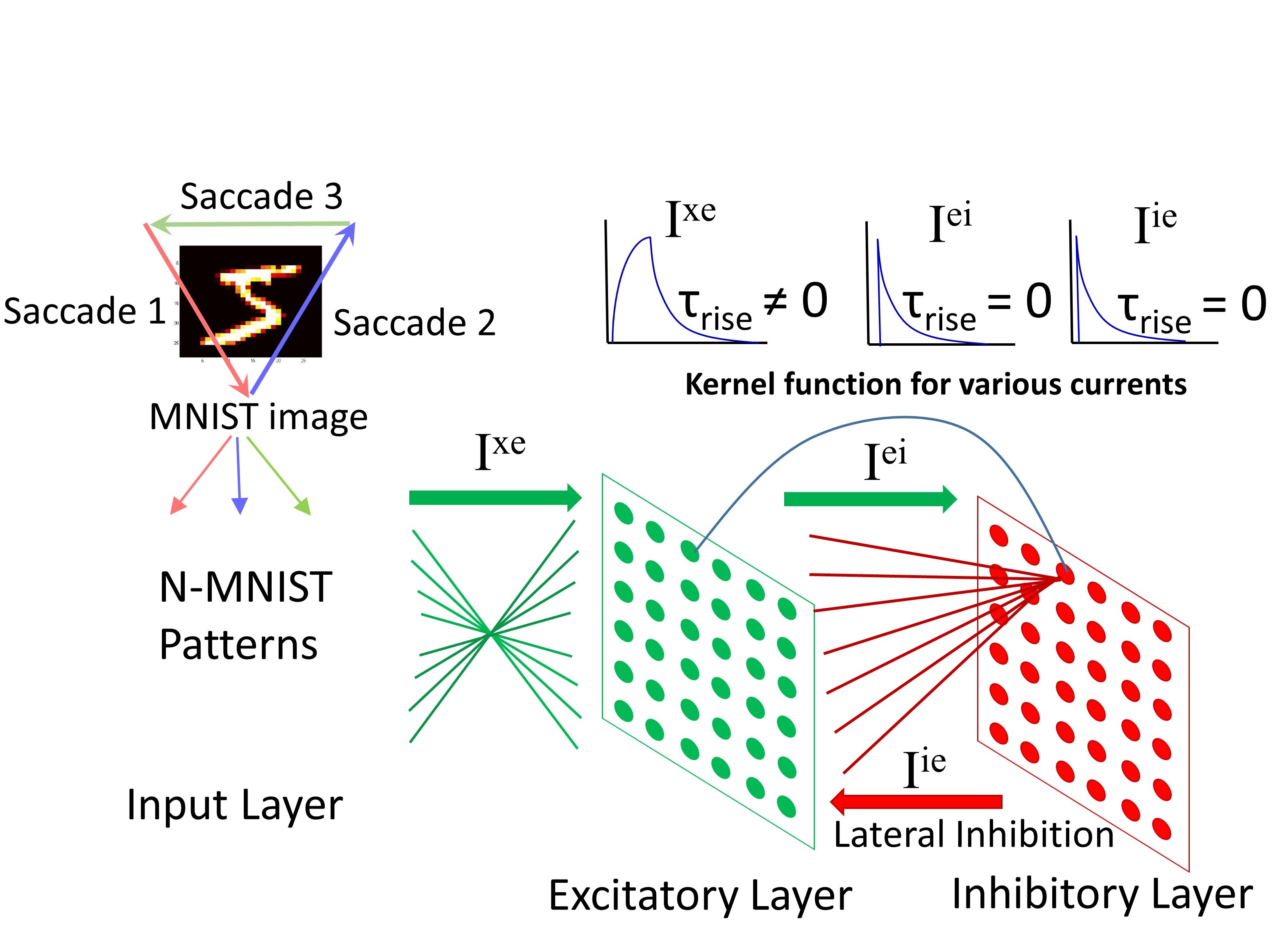}
	
	\caption{\color{Gray} \textbf{Spiking Neural Network Architecture.}. (Modified version of Figure 2 in \cite{Iyer:2017}) The N-MNIST patterns, each pattern representing a \emph{saccade} project to the excitatory layer. There is an all-to-all connection from the input to excitatory layer. The excitatory layer has a one-to-one connection with the inhibitory layer. Upon firing, an excitatory neuron activates its corresponding inhibitory neuron, which in turn inhibits all excitatory neurons except the one it received excitatory connections from. Top right: Kernel functions for the currents in the network - $I_{xe}$ has a gradual rise in current followed by a gradual fall. The other two currents, $I_{ei}$ and $I_{ie}$ have an instantaneous rise in current followed by a gradual fall.}
	
	\label{fig:NA} % \label works only AFTER \caption within figure environment
	
\end{figure}

The input layer contains $34 \times 34$ neurons (one neuron per image pixel in N-MNIST). Each input neuron projects to all neurons in the \emph{excitatory layer} with weights $W^{xe}$. The \emph{excitatory layer} has $N_e$ neurons which have a one-to-one connectivity with $N_i$ neurons in the \emph{inhibitory layer}. Note that $N_i = N_e$.  When a neuron spikes in the \emph{excitatory layer} it will activate the corresponding neuron in the \emph{inhibitory layer}. Each inhibitory neuron inhibits all neurons in the \emph{excitatory layer} except the one that it has afferent excitatory connection with. The net effect is lateral inhibition. 

The system architecture is shown in Figure \ref{fig:NA}. 

\subsection{Network Dynamics} \label{netDy}

The network is made up of leaky integrate and fire (LIF) neurons. The membrane potential $V$ of neuron $j$ at time $t$ in the \emph{excitatory layer} is given as follows: 

\begin{align} \label{IFeqn}
\tau_M \frac{dV_j}{dt} &= - (V_j-V_{leak}) + RI_j(t) \notag \\
V_j(t) \geq V_{th} &\implies t \in s^e_j; V_j(t) \leftarrow V_{reset} 
\end{align}

where $\tau_M$ is the membrane time constant in $ms$, $V_{th}$ is the threshold voltage in $mV$, $V_{leak}$ is the leak voltage in $mV$ and $I_j(t)$ is the total current in $nA$ that is input into the neuron $j$. $s^e_j$ is the spike train of neuron $j$ in the \emph{excitatory layer}. It is a set containing all the spike times of neuron $j$. The membrane potential in the \emph{inhibitory layer} is likewise calculated, and $s^i_j(t)$ is the spike train of neuron $j$ in the \emph{inhibitory layer}. After a spike is fired, the membrane voltage of the neuron is set to $V_{reset}$ and the neuron is refractory for the absolute refractory period $t_{ref}$, before it can fire again. 
%but then if you write like that, how to put it in th equation? Now ask - how to put it in the 
The net current into a neuron $j$ in the \emph{excitatory layer} is the sum of currents from the input and inhibitory layers. 

\begin{equation} \label{eqn:SNNcurTotal}
I_j(t) = I^{xe}_j(t) +  I^{ie}_j(t)
\end{equation}

where $I^{xe}_j(t)$ is the excitatory current from the input layer, and $I^{ie}_j(t)$ is the inhibitory current from the \emph{inhibitory layer} to neuron $j$ in the \emph{excitatory layer}. 

The current into each neuron $j$ in the \emph{inhibitory layer} is received from the \emph{excitatory layer} alone. 

The current $I^{xe}_j(t)$ into neuron $j$ is calculated as: 

\begin{align} \label{eqn:currentEqns}
I^{xe}_j(t) &= A^{xe} \sum\limits_k g^{xe}_k W^{xe}_{jk} \notag \\
g^{xe}_k(t) &= K^{xe}(t) * s^e_k(t) \notag \\
K^{xe}(t) &= (e^{\frac{-t}{\tau^{xe}_{fall}}} - e^{\frac{-t}{\tau^{xe}_{rise}}}) u(t)
\end{align}

% if we change the defininition, then all thes need to be changed. it becomes a little different and time consuming... necessary? 
where $u$ is the Heaviside function and $K^{xe}$ is the input kernel function. $g^{xe}(t)$ is the input convolved with the kernel. $s^e_k(t)$ is the input spike train of neuron $k$. $*$ is the convolution operator. The currents $I^{ie}$ and $I^{ei}$ are similarly calculated, using weights $W^{ei}_{jk}$ and $W^{ie}_{jk}$ respectively, and $s^e_l(t)$ and $s^i_m(t)$ are the spike trains of neurons $l$ and $m$ in the \emph{excitatory} and \emph{inhibitory} layers respectively. As $A^{xe}$ is increased (see Section \ref{inpenc}), it is possible to have a few neurons spike simultaneously and inhibit one another. To prevent that, the kernel function for $I^{xe}$ has an alpha waveform with gradual rise in current. The other current ($I^{ie}$, $I^{ei}$) kernels have exponential waveforms with instantaneous rise in current similar to the original method by \cite{Diehl:2015} (see Figure \ref{fig:NA}, Top right). Therefore the time constants for the rise in current, $\tau^{xe}_{rise} > 0$ but $\tau^{ie}_{rise}, \tau^{ei}_{rise} = 0$. 

\subsection{Learning} \label{SNNlearn}

The learning function follows from \cite{Diehl:2015}. When there is a postsynaptic spike, the synaptic weight update $\Delta w$ is:  

\begin{equation} \label{eqn:SNNlearn}
\Delta w = \eta (x_{pre} - x_{tar}) (w_{max} - w)^\mu
\end{equation}

where $x_{pre}$ is the presynaptic trace, $x_{tar}$ is the target value of the presynaptic trace at the moment of postsynaptic spike, $\eta$ is the learning rate, $w_{max}$ is the maximum weight, and $\mu$ determines the dependence on the previous weight. See \cite{Diehl:2015} for more details. 

When a presynaptic spike arrives at the synapse, the presynaptic trace, $x_{pre}$ is increased by $\Delta {x_{pre}}$, and decays exponentially with the time constant $\tau_{x_{pre}}$. 

\subsection{Threshold adaptation} \label{sec:thrAdapt}

The threshold adaptation mechanism used here is identical to that employed by \cite{Diehl:2015}. In order to prevent any single neuron in the excitatory layer from dominating the response pattern, it is desirable that all neurons have similar firing rates at the end of training. Therefore, the neuron's firing threshold $V_{th}$ is adapted as follows: 

\begin{equation} \label{eqn:SNNvth}
V_{th} = v_{thresh} + \theta
\end{equation}

$V_{th}$ has two components, a constant $v_{thresh}$ and a variable component, $\theta$. $\theta$ is increased by $\Delta \theta$ every time a neuron fires, and decays exponentially. Therefore if a neuron spikes more, its threshold is higher, requiring more input for the neuron to spike. 

\subsection{Pattern Presentation} \label{inpenc}

If for any pattern presentation there is no output spike, $A^{xe}$ is increased by $\Delta A^{xe}$ and the pattern is presented again. This is repeated till there is an output spike. 

\subsection{Neuron Label Assignment}

Once the training is done, the training patterns are presented again to the learnt system. Each neuron is assigned to the class that it most strongly responds to. This neuron assignment is used in calculating the classification accuracy. Note that class labels are only used in this step, and not for training. 

\subsection{Parameters}\label{Params}

The values of most parameters in this SNN follow \cite{Diehl:2015}. These include $V_{rest}$, $v_{thresh}$ and $V_{reset}$ in the \emph{excitatory} and \emph{inhibitory} layers. Since we present each pattern one after another, the presentation time is $105$ ms, equivalent to the time taken for one saccade in the N-MNIST dataset. As presynaptic spike rates vary throughout pattern presentation, the output neuron must spike only at the end of the presentation (see \cite{Iyer:2017} for more details). Therefore, $\tau_M$, the membrane time constant of each excitatory neuron is adjusted such that there is only one output spike (see \cite{Iyer:2017} for additional information) occurring towards the end of pattern presentation. After each pattern presentation, all values except $W_{xe}$ and $\theta_e$ are reset. 

For the \emph{Design Space Explorations} (See Section \ref{sec:DSE}), time constant of presynaptic trace - $\tau_{xpre}$, learning rate - $\eta$ and amplitude of threshold adaptation - $\Delta \theta$ are adjusted accordingly. 

In the sections that follow we describe the experiments that use the SNN described above. 

\section{Experiment: Design Space Exploration in SNN to explore temporal and rate-based STDP regimes} \label{sec:DSE}

\begin{figure}[ht] %s state preferences regarding figure placement here
	
	% use to correct figure counter if necessary
	%\renewcommand{\thefigure}{2}
	
	\includegraphics[width=\textwidth]{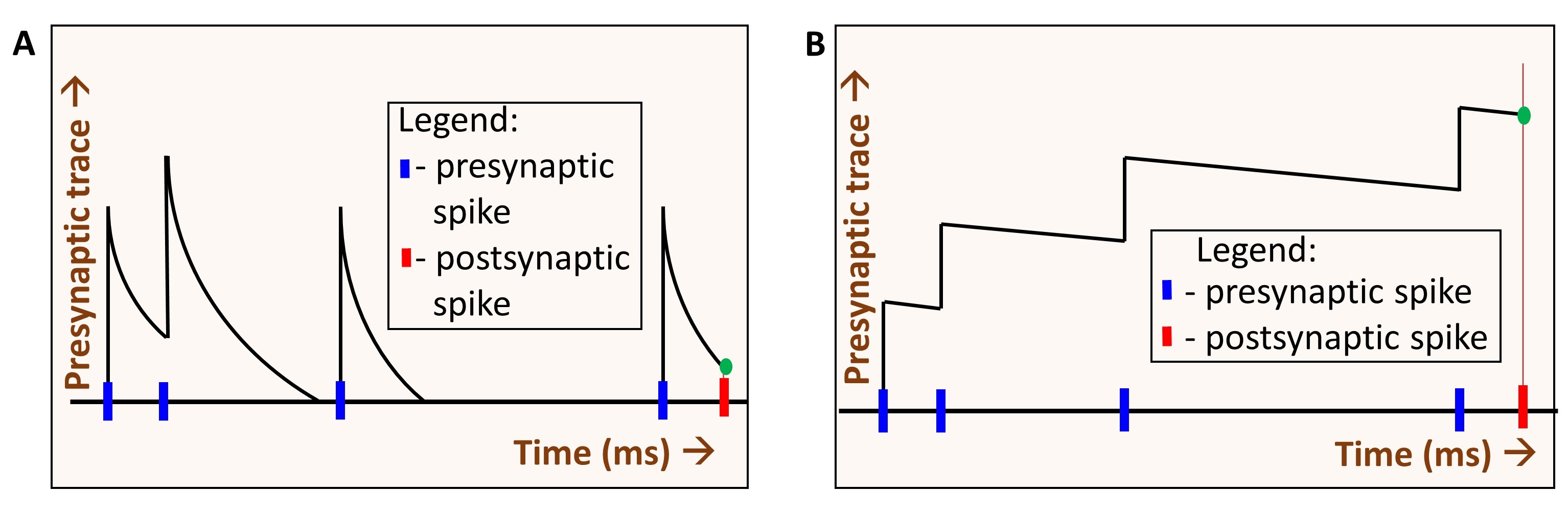}
	
	\caption{\color{Gray} \textbf{Two regimes of STDP operation.} By varying the decay time constant of the presynaptic spike trace $\tau_{xpre}$ we get two regimes of STDP operation, (A) When $\tau_{xpre}$ is low, the presynaptic trace ($x_{pre}$) decays quickly. The value $x_{pre}$ at the time of the postsynaptic spike (green dot) depends on the time difference between the pre- and post-synaptic spikes. In this regime, spikes that occurred much earlier than the postsynaptic spike time have no impact on learning. (B) When $\tau_{xpre}$ is high, $x_{pre}$ decays slowly. At the time of postsynaptic spike, $x_{pre}$ (green dot) depends on the number of spikes (i.e. spike-count rate) alone. Precise presynaptic spike times do not have much impact on learning. }
	
	\label{fig:STDP} % \label works only AFTER \caption within figure environment
	
\end{figure}

Spike-timing dependent plasticity (STDP) is one learning rule commonly used in SNNs for unsupervised learning. Generally in STDP, weight updates are based on the precise difference between pre and postsynaptic spike times. However, in a regime whereby the single post-synaptic spike is constrained to be near the end of a pattern, STDP can operate in two different modes, where weight updates depend mostly on time averaged spike rates or time difference between the more recent presynaptic spikes and the single postsynaptic spikes. These different modes due to presynaptic time constant ($\tau_{xpre}$) are illustrated in Figure \ref{fig:STDP}. Note that the shape of the PSTH of the input spike train is low in the beginning, peaks in the middle, and is low again in the end. Although there is only one postynaptic spike, it happens mostly after the peak of the PSTH, and therefore, captures the most important input information within the time window dictated by the presynaptic time constant $\tau_{xpre}$. 

The hypothesis for our experiment is – if precise spike-times were important in N-MNIST, our SNN would perform poorly in the rate-based regime. This is because information on time averaged spike rates alone would be inadequate for classification, as the precise timing of spikes would be necessary. We therefore compare the performance of N-MNIST on the rate based and time based regimes by systematically varying several parameters – as listed below. 

\begin{itemize}
	\item $\tau_{xpre}$, Time constant of presynaptic trace (see Section \ref{SNNlearn}, last line)- As shown in Figure \ref{fig:STDP}, varying $\tau_{xpre}$ will result in the rate-based or time based regimes of STDP on the two extremes of a continuum. 
	\item $\eta$, Learning rate - Higher values of $\tau_{xpre}$ would result in higher values of the presynaptic trace, $x_{pre}$ as individual spike traces would decay slowly. This results in an accumulation of individual spike traces over time. This, in turn, would lead to higher weight updates (see the learning rule (Equation \ref{eqn:SNNlearn}) in Section \ref{SNNlearn}). To ensure that results are not biased due to more learning in the system, we also vary $\eta$. 
	\item $\Delta \theta$, Amplitude of threshold adaptation (see Section \ref{sec:thrAdapt}) - Threshold adaptation is done to prevent some neurons from dominating the learning and distributing the receptive field of input patterns over all neurons. However, if threshold adaptation occurs very slowly compared to the learning rate, this purpose will not be served. If, on the other hand, the threshold of a neuron is increased very quickly before it even learns, then during training, no useful learning will take place. We therefore change $\Delta \theta$ along with $\eta$. 
\end{itemize} 

\begin{figure}[ht] %s state preferences regarding figure placement here
	
	% use to correct figure counter if necessary
	%\renewcommand{\thefigure}{2}
	
	\includegraphics[width=10cm]{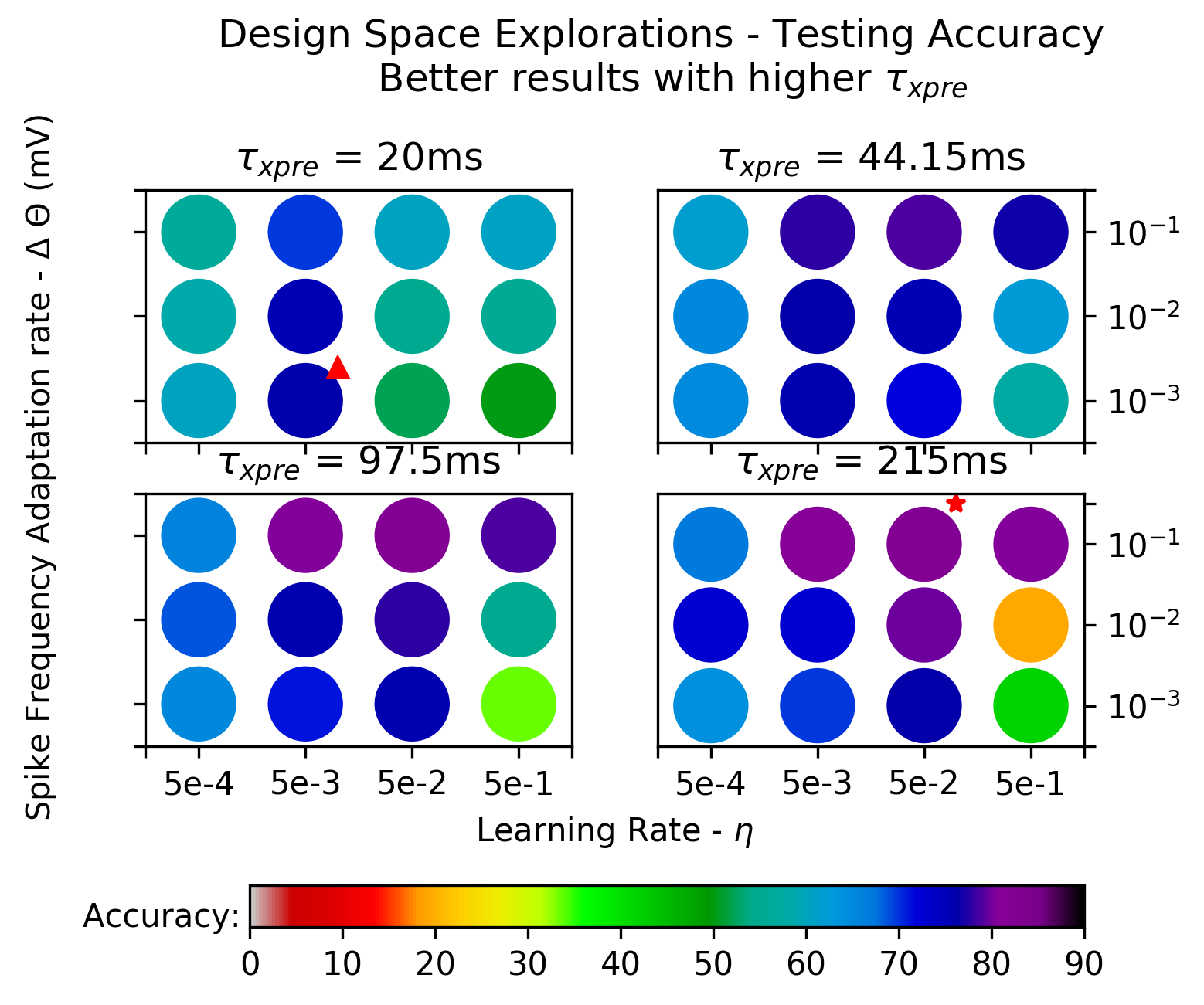}
	
	\caption{\color{Gray} \textbf{Results on \emph{log scale}.}Design Space Exploration results, all parameters are plotted on a $log$ scale, trained over one epoch. Results progressively get better as $\tau_{xpre}$ increases. The best overall result (82.46\%) is marked by a red star in the top right. The best overall results, and a larger number of better results ($>80$\%) occur at higher values of $\tau_{xpre}$. The best result for the low $\tau_{xpre}$ (= 20 ms) is marked by a red triangle on the top right. }
	
	\label{fig:results} % \label works only AFTER \caption within figure environment
	
\end{figure}

\begin{figure}[ht] %s state preferences regarding figure placement here
	
	% use to correct figure counter if necessary
	%\renewcommand{\thefigure}{2}
	
	\includegraphics[width=10cm]{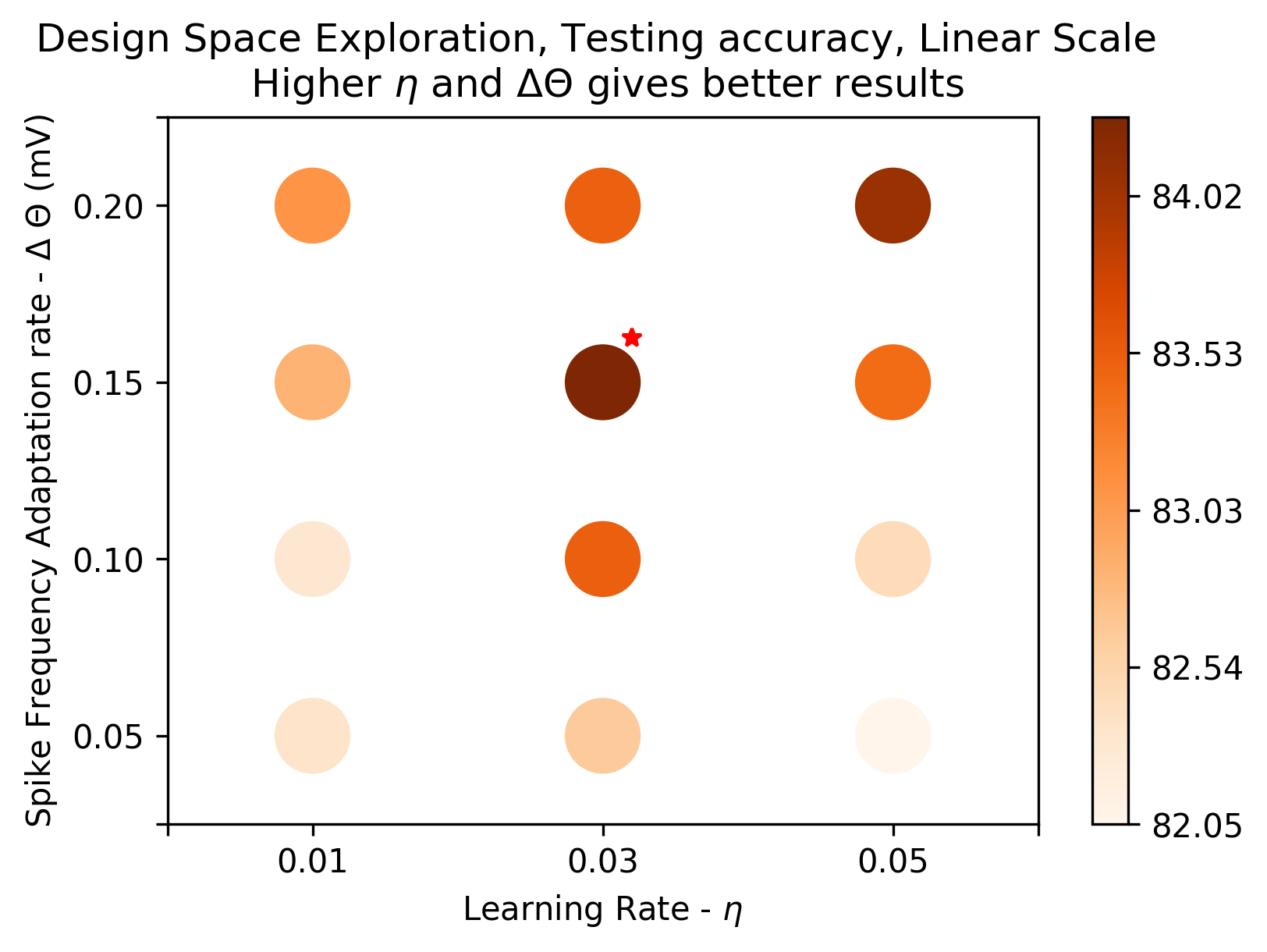}
	
	\caption{\color{Gray} \textbf{Results on \emph{linear scale}.} The best result in the \emph{log scale}, (Figure \ref{fig:results}, $\tau_{xpre} = 215 ms, \eta = 0.05, \Delta \theta = 0.1 mV$) was taken. $\tau_{xpre}$ was fixed at $215$ ms and values of the other parameters ($\Delta \theta$ and $\mu$) around the best result were plotted on linear scale to determine if there were better results around the parameter space. Training was done over one epoch. We see that generally results get better as learning rate ($\eta$) gets higher and spike frequency adaptation rate ($\Delta \theta$) gets higher. The best results are marked with a red star on the top right. We get a slight improvement in accuracy, $84.51\%$. }
	
	\label{fig:resultsZoom} % \label works only AFTER \caption within figure environment
	
\end{figure}

We use these three parameters for the DSE. All parameters are varied on a logarithmic scale to ensure that we cover all possible ranges of activity. The values of $\eta$ used are $\{0.0005, 0.005, 0.05, 0.5\}$. The values of $\Delta \theta$ used are $\{0.001mV, 0.01mV, 0.1mV\}$. $\tau_{xpre}$ values range from 20 ms (the initial value used in \cite{Diehl:2015}) to more than $2 \times$ the pattern presentation time, 215ms (this is the value used in \cite{Iyer:2017}) and have intermediate values on a logarithmic scale, $\{20ms, 44.15ms, 97.5ms, 215ms\}$. The results are shown in Figure \ref{fig:results}. As results are on the \emph{logarithmic} scale, there may be intermediate parameter values that give better results. To check this, we repeat the experiments for parameters  around the vicinity of the best results seen in Figure \ref{fig:results} (i.e. \emph{logarithmic} scale) and these are then plotted in \emph{linear} scale. The results are given in Figure \ref{fig:resultsZoom}. 

% From Figure X we see that as $\tau_{xpre}$ increases, the best results over $\eta$ and $\Delta \Theta$ get better,  the number of values, and best values over different values of $\Delta \theta$ and $\eta$ increase. ## Put this in the figure caption. 

As we can see in Figure \ref{fig:results}, an increase in $\tau_{xpre}$ clearly denotes better overall results. There was a slight improvement in results, when we tried parameter values near the best result, as shown in Figure \ref{fig:resultsZoom}. This experiment shows that rate based regime of STDP is able to give overall best results, denoting that time averaged spike rates are sufficient to characterize N-MNIST and achieve good accuracy. 

As training was done only on one epoch, a natural question that arises is: do lower values of $\tau_{xpre}$ yield bad results just because they were not sufficiently trained? What if they were trained for multiple epochs? We explore training over multiple epochs in the next section. 

\subsection{Training over multiple Epochs}

In the results presented earlier (Figures \ref{fig:results} and \ref{fig:resultsZoom}), training was only done on one epoch. We have chosen the best results after one epoch for $\tau_{xpre} = 20 ms$ and $\tau_{xpre} = 215 ms$ which are: 

\begin{itemize}
	\item \textbf{Best result for $20 ms$ -} Accuracy: $76.25\%$ for parameter set $\{\tau_{xpre} = 20 ms$, $\Delta \theta = 0.001 mV$, $\mu = 0.005 ms\}$, and 
	\item \textbf{Best result for $215 ms$ -} Accuracy: $82.46\%$ for parameter set $\{\tau_{xpre} = 215 ms$, $\Delta \theta = 0.01 mV$, $\mu = 0.05 ms\}$. 
\end{itemize}

We have run the SNN with these two parameter sets for 4 epochs. 

\begin{figure}[ht] %s state preferences regarding figure placement here
	
	% use to correct figure counter if necessary
	%\renewcommand{\thefigure}{2}
	
	\includegraphics[width=\textwidth]{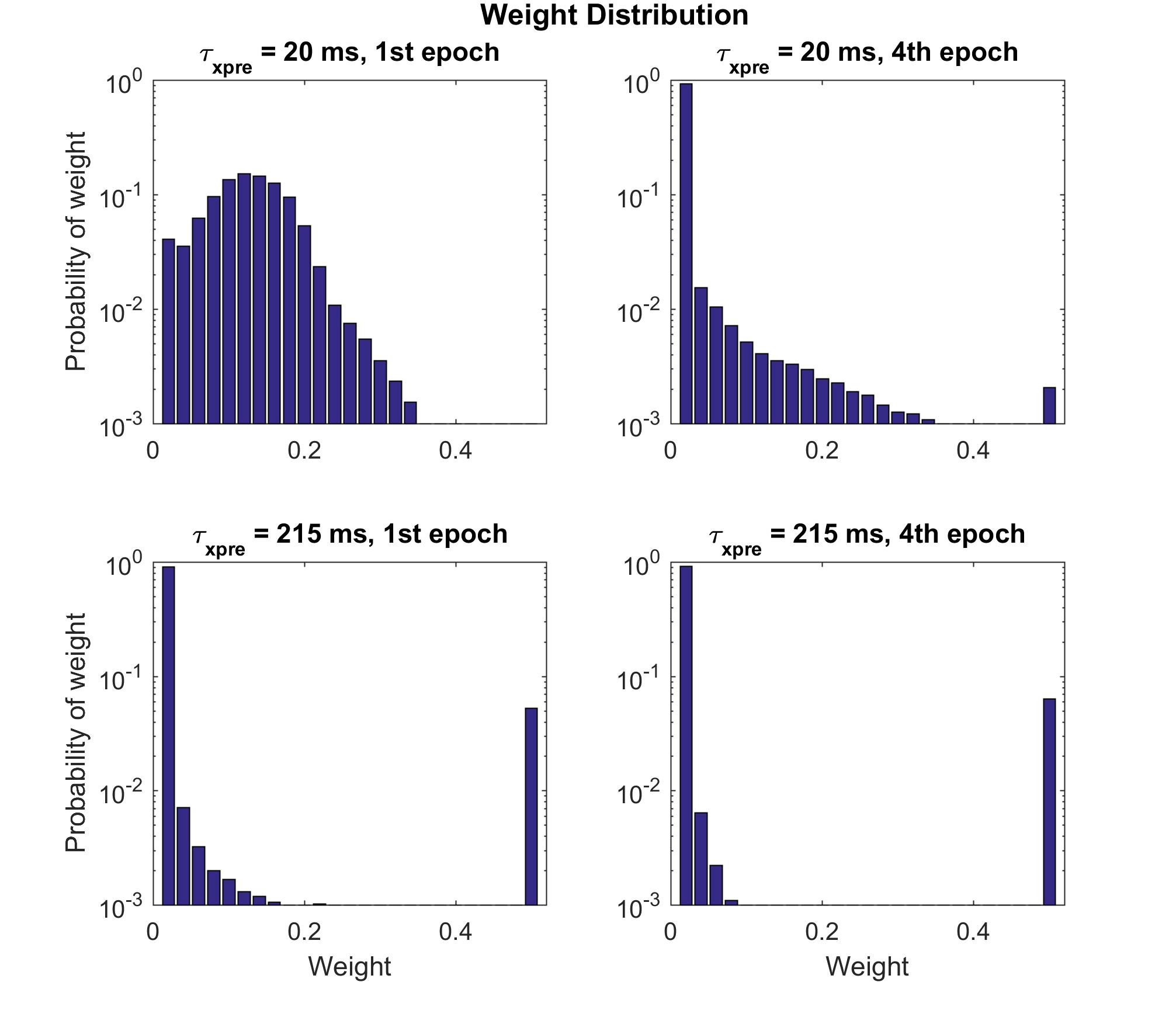}
	
	\caption{\color{Gray} \textbf{Weight distribution} plotted after training for one epoch (left) and training for 4 epochs (right). This is plotted for the best parameter values for $\tau_{xpre} = 20$ ms (Threshold adaptation rate, $\Delta \Theta = 0.001 mV$, Learning Rate, $\mu = 0.005$) and $\tau_{xpre} = 200$ ms (Threshold adaptation rate, $\Delta \Theta = 0.01 mV,$ Learning Rate, $\mu = 0.05$). After running more epochs, the networks become more stable. }
	
	\label{fig:weightDist} % \label works only AFTER \caption within figure environment
	
\end{figure}

\begin{figure}[ht] %s state preferences regarding figure placement here
	
	% use to correct figure counter if necessary
	%\renewcommand{\thefigure}{2}
	
	\includegraphics[width=\textwidth]{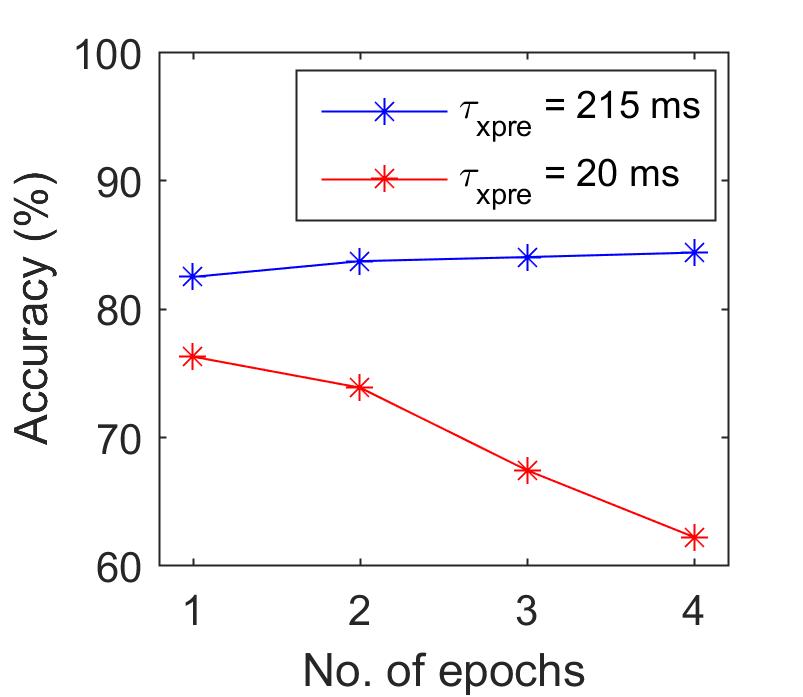}
	
	\caption{\color{Gray} \textbf{Testing accuracy over epochs.} Testing accuracy is plotted as a function of number of training epochs. This is plotted for the best parameter values for $\tau_{xpre} = 215$ ms (blue) (Threshold adaptation rate, $\Delta \Theta = 0.01 mV,$ Learning Rate, $\mu = 0.05$) and $\tau_{xpre} = 20$ ms (red) (Threshold adaptation rate, $\Delta \Theta = 0.001 mV$, Learning Rate, $\mu = 0.005$). As can be seen, results improve over time when $\tau_{xpre}$ is high, but actually decrease over time when $\tau_{xpre}$ is low. }
	
	\label{fig:accOverEpochs} % \label works only AFTER \caption within figure environment
	
\end{figure}

\begin{figure}[ht] %s state preferences regarding figure placement here
	
	% use to correct figure counter if necessary
	%\renewcommand{\thefigure}{2}
	
	\includegraphics[width=\textwidth]{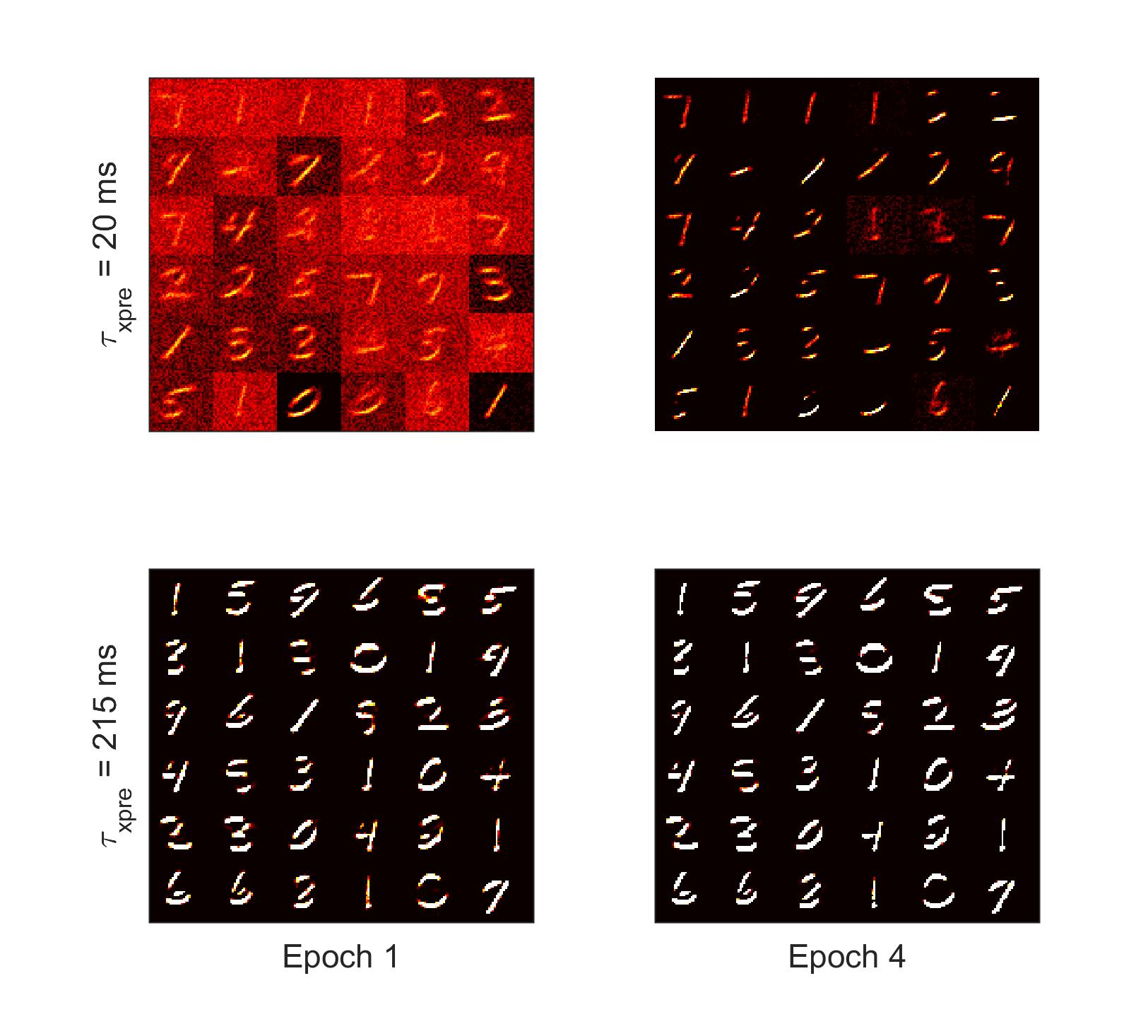}
	
	\caption{\color{Gray} \textbf{Weights over epochs.} Weights obtained after training on one epoch (left) and 4 epochs (right) for $\tau_{xpre} = 20$ ms (top row) and $\tau_{xpre} = 215$ ms (bottom row) are plotted. The weight for each neuron is plotted on a 34 $\times$ 34 grid. For the low $\tau_{xpre}$ (= 20 ms) case, while the digits can be vaguely discerned after training on one epoch, after training for 4 epochs, only part of the digit is visible. (For e.g. Digit ``2'' on the top right corner and digit ``5'' on the bottom left corner.) On the other hand, for the high $\tau_{xpre}$ case (= 215 ms) digits become slightly more discriminative when training on more epochs. (For e.g. digit ``3'' third row far right. }
	
	\label{fig:WoverEpochs} % \label works only AFTER \caption within figure environment
	
\end{figure}

In both cases, as the network runs for 4 epochs, the weights tend to a more bimodal distribution, where weights congregate towards the maximum and minimum weight values (Figure \ref{fig:weightDist}). This indicates network convergence (\cite{Kempter:1999}).  However, although there is an improvement in accuracy for the high value of $\tau_{xpre}$, this is not the case for the low value of $\tau_{xpre}$ (Figure \ref{fig:accOverEpochs}). This is because for high values of $\tau_{xpre}$ the weights get crisper and more discriminative, but for low values of $\tau_{xpre}$ only part of the image is being learnt \ref{fig:WoverEpochs}. This is possibly because only input spikes that  occur shortly before the postsynaptic spike induce long term potentiation (LTP) in a time based STDP regime (Figure \ref{fig:STDP}).   

This experiment indicates that it is not inadequate learning that led to worse accuracy for low values of $\tau_{xpre}$. In fact with more training, we see that only rate-based STDP has an improvement in accuracy, while being more trained on a short time window of the pattern (albeit the time window with the most inputs generally) using time based STDP results in poorer accuracy. This is a direct consequence of the bimodal distribution of weights with more training \ref{fig:weightDist}(Right): synaptic connections corresponding to inputs within that time window undergo LTP, while the others undergo long term depotentiation (LTD).  

We then train the SNN on 3 epochs with the parameters that get the best results on one epoch in Figure \ref{fig:resultsZoom} to get an accuracy of 89.87\%. 

Finally we repeated this procedure for training 6 separate 400 neuron networks - each of the 3 saccades with both ON and OFF polarities. For each test pattern, all 6 networks gave a class prediction and we took a majority vote. Using this method, an accuracy of 91.78\% was obtained. 

\subsection{STDP with fixed Postsynaptic Spike}

Earlier we noted that rate based STDP regime yields the best accuracy results indicating that presynaptic spike times do not affect the accuracy. If learning is dependent on purely spike rates alone, we postulate that the precise timing of postsynaptic spikes should not affect the accuracy either. So if we fix the postsynaptic spike to occur at a certain time for every pattern, there should not be a fall in accuracy. 

In this experiment we train the system using the parameters for the best results seen in the previous experiment (Figure \ref{fig:resultsZoom}: $\Delta \theta = 0.2 mV, \eta = 0.05, \tau_{xpre} = 215ms$), and record the postsynaptic spike time for each pattern. We then find the average of the postsynaptic spike time over all patterns, $t^*$. 

We re-start and repeat the training fixing the postsynaptic time to be $t^*$. As we have a Winner-Take-All network, and the neuron that is the first to spike wins, we do not enforce the neuron to spike at time $t^*$, but the network learns \emph{as if} the spike occurs at $t^*$. Therefore, we take the presynaptic spike traces at time $t^*$ to calculate the weight updates. 

The accuracy for this experiment is $84.10\%$ after one epoch. This is even better than the best accuracy results in the DSE experiment \ref{sec:DSE} which is $82.46\%$. The high accuracy indicates that performance is not dependent on the precise timing of the postsynaptic spike either.  

\section{Experiment: Spike Rate Dependent STDP} \label{sec:popRate}

In the previous experiments, we examined the performance of a simple ANN and the SNN (both rate based and time based) on the N-MNIST dataset. Both the ANN and the rate based SNN use time-averaged firing rates (Figure \ref{fig:rate}A) for classification. In this final experiment we examine the effect of instantaneous population rate (Figure \ref{fig:rate}B) on performance. 

We note that events recorded by the ATIS sensor are relatively sparse at the beginning and end of a saccade. Most events occur in the middle of a saccade. So we hypothesize that the middle of the saccade is the time period where the information is the most abundant. Events that happened at the beginning and end of the saccade could be regarded as noise. From this, we hypothesize that by 1. using an \emph{engineered} STDP function – i.e. an STDP function that is based on the peristimulus time histogram (PSTH) of the training data, and 2. fixing the postsynaptic spike time at the end of the pattern presentation, we will not experience a decrease in performance.  
% maybe you want to ask comments/ 
Such an STDP function is completely independent of the pre and postsynaptic spike time differences, and is governed by the instantaneous population spike rates alone. If the above hypothesis is correct, then spike times of individual neurons are unnecessary. Instantaneous population spike rates adequately characterize the dataset. 

The STDP function is created as follows. Each pattern $p$ can be represented as a set of spike trains, with one spike train for each pixel. The spike train for pattern $p$ and pixel $x$ is represented as $s^{x, p} = \{t^{x, p}_1, t^{x, p}_2, ... t^{x, p}_n \}$, where each of the elements represents the time at which the corresponding event occurred. Note that $t_1, ... , t_n$ are in the range $[0,105]$ ms, (first saccade) and are all $ON$ events.  

The total number of events that occurred over all patterns $p$ at all pixels $x$ at the instantaneous time between $t$ and $\Delta t$ is: 

\begin{equation} \label{eqn:popTotal}
H'(t) = \sum_p \sum_x a^{x, p}_i
\end{equation}

\begin{equation} \label{eqn:popTempvar}
a^{x, p}_i = \left\{ \begin{array}{ll} 1 , & t \leq t^{x, p}_i \leq  \Delta t, t^{x, p}_i \in s^{x, p};\\
0, & \textrm{otherwise}.
\end{array}\right.
\end{equation} 

\begin{equation} \label{eqn:popFrac}
H(t) = \frac{H'(t)}{N_{patterns}} 
\end{equation}

$H(t)$ is then scaled and biased as follows: 

\begin{equation}
h(t) = a H(t) + b
\end{equation} 

Parameters $a$ and $b$ are chosen so that the STDP function $h(t)$ fulfills the following conditions: 

\begin{itemize}
	\item The area of LTD is greater than the area of LTP - this is to ensure network stability (\cite{Song:2000}).  
	\item The weight updates are of similar magnitude to that of the learning rule described in Section \ref{SNNlearn}. This is determined empirically using the first few patterns so as to determine $a$ and $b$, so that learning rate would not be the varying factor in the classification accuracy obtained in experiments.
\end{itemize}

\begin{figure}[ht] %s state preferences regarding figure placement here
	
	% use to correct figure counter if necessary
	%\renewcommand{\thefigure}{2}
	
	\includegraphics[width=\textwidth]{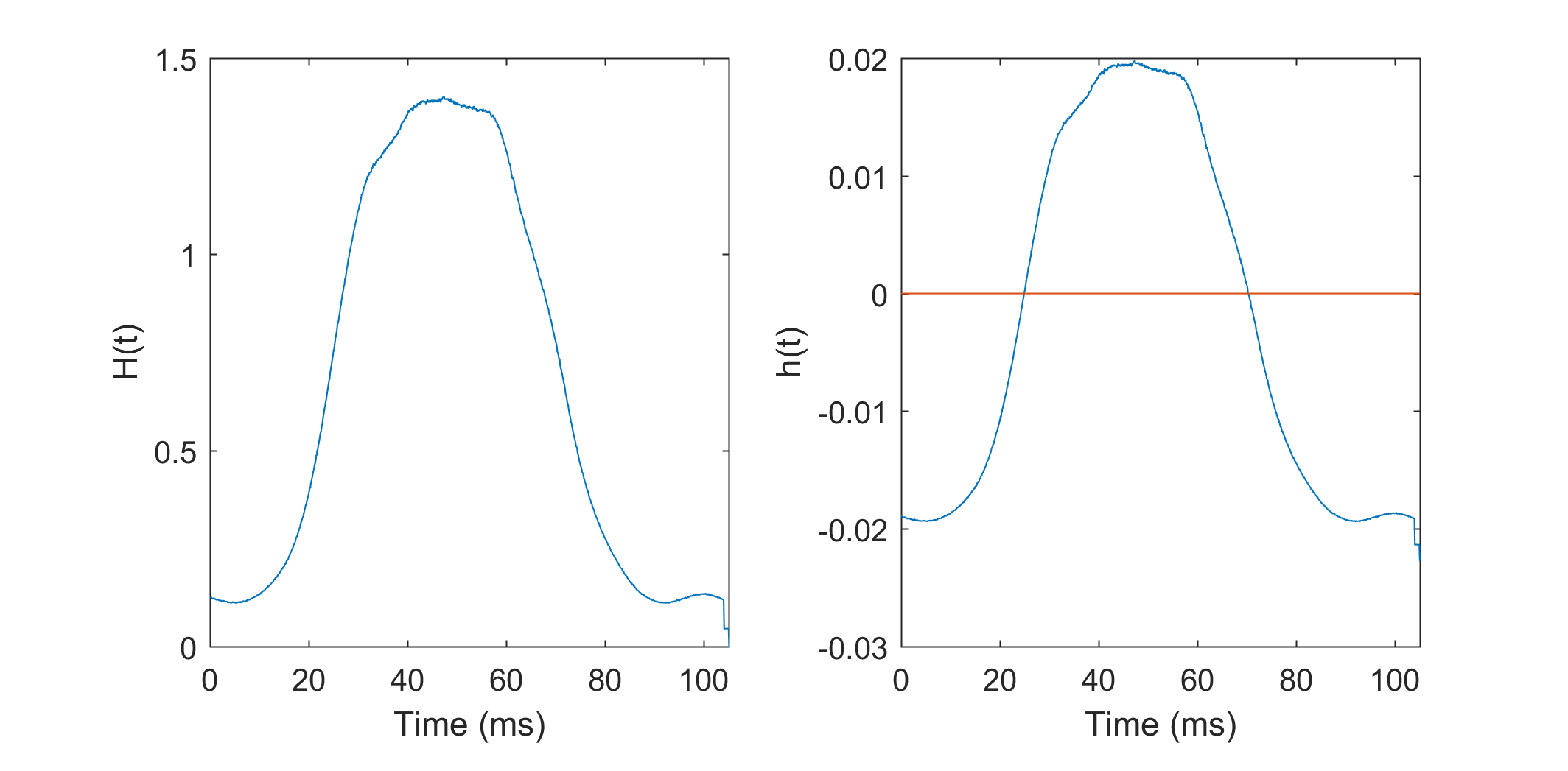}
	
	\caption{\color{Gray} \textbf{STDP function derived from the training data.} \emph{Left:} $H(t)$ is the average number of presynaptic neurons spiking at instantaenous time $t$ in a pattern. The function $H(t)$ derived from all the training patterns is given in the figure. \emph{Right:} $h(t)$ is the STDP curve obtained after scaling and biasing $H(t)$ to ensure stability and to preserve the learning dynamics described in the previous sections. }
	
	\label{fig:popRateFns} % \label works only AFTER \caption within figure environment
	
\end{figure}

The function $H(t)$ that we derived from the N-MNIST training data and the corresponding STDP function $h(t)$ that we obtained are given in Figure \ref{fig:popRateFns}. 

From the STDP function $h(t)$ we calculate the presynaptic trace $x_{pre}$ for a pattern $p$ as follows: 

\begin{equation} \label{eqn:popxpre}
x_{pre}^{x, p} = \left\{ \begin{array}{ll} \sum_{i=1}^{i=n} h(t^{x, p}_i) , & n > 0\\ 
-x_{tar}, & \textrm{otherwise}.
\end{array}\right.
\end{equation}

So when a neuron fires one or more spikes, the resultant value $x_{pre}$ is the sum of the values of $h(t)$ for all time instances $t$ where spikes occurred. Equation \ref{eqn:popxpre} also has a depression component. When a neuron does not spike at all, there is a LTD of $-x_{tar}$. This is similar to Section \ref{eqn:SNNlearn}, equation \ref{eqn:SNNlearn}, where, in the absence of spikes, a neuron gets depressed by the same amount $-x_{tar}$. The LTD component is introduced in equation \ref{eqn:popxpre} to keep the learning dynamics similar to that of Section \ref{SNNlearn}. This addition of depression does not negate the purpose of this experiment - the STDP curve is still dependent on the presynaptic spike rate. 

The learning rule is similar to that of equation \ref{eqn:SNNlearn} in Section \ref{SNNlearn}: 

\begin{equation} \label{eqn:popLearn}
\Delta w = \eta x_{pre} (w_{max} - w)^\mu
\end{equation}

where $x_{pre}$ is the presynaptic trace, $\eta$ is the learning rate, $w_{max}$ is the maximum weight, and $\mu$ determines the dependence on the previous weight.

We trained the SNN using this learning rule above for one epoch, and we obtained an accuracy of $85.45\%$. Good performance was obtained on an PSTH derived STDP function. Postsynaptic spike time was also fixed. This indicates that precise time differences between pre and postsynaptic times are not necessary to classify the N-MNIST dataset. 

From the STDP function $h(t)$ we calculate the presynaptic trace $x_{pre}$ for a pattern $p$ as follows: 

\begin{equation} \label{eqn:popxpre}
x_{pre}^{x, p} = \left\{ \begin{array}{ll} \sum_{i=1}^{i=n} h(t^{x, p}_i) , & n > 0\\ 
-x_{tar}, & \textrm{otherwise}.
\end{array}\right.
\end{equation}

So when a neuron fires one or more spikes, the resultant value $x_{pre}$ is the sum of the values of $h(t)$ for all time instances $t$ where spikes occurred. Equation \ref{eqn:popxpre} also has a depression component. When a neuron does not spike at all, there is a LTD of $-x_{tar}$. This is similar to Section \ref{eqn:SNNlearn}, equation \ref{eqn:SNNlearn}, where, in the absence of spikes, a neuron gets depressed by the same amount $-x_{tar}$. The LTD component is introduced in equation \ref{eqn:popxpre} to keep the learning dynamics similar to that of Section \ref{SNNlearn}. This addition of depression does not negate the purpose of this experiment - the STDP curve is still dependent on the presynaptic spike rate. 

The learning rule is similar to that of equation \ref{eqn:SNNlearn} in Section \ref{SNNlearn}: 

\begin{equation} \label{eqn:popLearn}
\Delta w = \eta x_{pre} (w_{max} - w)^\mu
\end{equation}

where $x_{pre}$ is the presynaptic trace, $\eta$ is the learning rate, $w_{max}$ is the maximum weight, and $\mu$ determines the dependence on the previous weight.

We trained the SNN using this learning rule above for one epoch, and we obtained an accuracy of $85.45\%$. Good performance was obtained on an PSTH derived STDP function. Postsynaptic spike time was also fixed. This indicates that precise time differences between pre and postsynaptic times are not necessary to classify the N-MNIST dataset. 

\section{Conclusion}

We described two contributions in this paper. Firstly, we presented the first unsupervised SNN algorithm to be tested on N-MNIST, which achieved 91.78\% accuracy with all six saccades and polarities. Second and more importantly, we wanted to evaluate if neuromorphic datasets obtained from Computer Vision datasets with static images are discriminative in the time domain. We started the study with both N-MNIST and N-Caltech101, and performed several more experiments on N-MNIST alone to evaluate the same. The first experiment, training N-MNIST and N-Caltech101 with an ANN using backpropagation achieved 99.23\% and 78.01\% accuracies respectively. These were the best results obtained on the corresponding datasets to date. This in turn indicates that collapsing the patterns in time does not affect the performance. The other experiments were performed on the unsupervised SNN with saccade 1 and polarity = $ON$. Comparing rate and time based STDP, we noted that our SNN performs best on N-MNIST when it operates in a rate based manner, achieving the best performance of 89.87\% while training on 3 epochs, as compared to the best performance of 76.25\% using time based STDP (obtained after training on 1 epoch. Training on more epochs does not improve accuracy). We further showed that fixing the postsynaptic spike time does not affect the performance. Finally, we experimented on an instantaneous population rate based STDP function, and this achieved a performance of 85.45\%.  This shows that the instantaneous rate over a population of neurons fully characterizes the N-MNIST dataset. Collectively these experiments show that in the N-MNIST dataset, the precise timings of individual spikes are not critical for classification. 

We initially approached N-MNIST to devise a STDP algorithm for classifying neuromorphic data, and as a result we implemented the first unsupervised SNN algorithm for N-MNIST. However, explorations with N-MNIST showed that its features encoded are not discriminative in time. These results are confirmed in N-Caltech101 as well. In this section, we detail why this result is important, and discuss the possible next steps. We pose several questions: 1.) Why do we get these results? 2.) Why do we need a neuromorphic dataset that is discriminative in the time domain? 3.) What constitutes a good neuromorphic dataset? If N-MNIST is not suitable, then what is? This is a very important question in neuromorphic engineering.   

Why do we get these results? We get good results in the ANN (Section \ref{sec:ANN_TF}) and rate-based SNN (Section \ref{sec:DSE}) due to the nature of N-MNIST. We sum up the spikes in an N-MNIST saccade in two ways 1.) through collapsing the events in time as in Section \ref{sec:ANN_TF} or 2.) by a relatively non-leaky integration of spikes in Section \ref{sec:DSE}. Using both methods, we note that after summation we retain all the information in N-MNIST \ref{fig:sqImg}. This is possibly because of the static 2-dimensional nature of the underlying dataset (i.e. MNIST). Using the N-MNIST creation process of recording from the ATIS camera can at best reproduce the original MNIST dataset – there is no additional information over time. N-MNIST is less informative than MNIST, due to noise and gradations in the image introduced due to the moving camera. Noise is good, as the recordings from the camera make the dataset more realistic. Gradation in the image – i.e. high spike rate while recording certain parts of the image and low spike rates in other parts of the image – is an artefact introduced by the predefined and regular N-MNIST camera movements. Such gradations do occur in the real world. However, as our sensory neurons are able to detect and embody the statistics in the environment (\cite{Simoncelli:2001}, \cite{Geisler:2008}, \cite{Elder:2016}) the image gradations that are represented are probably those that are actually present in scenes. 

We get good results in the last experiment (Section \ref{sec:popRate}) due to an artefact in the N-MNIST dataset. The ATIS camera movements are clearly defined, regular, and all images are relatively similarly sized. Such regularity is not characteristic of retinal saccades, or any other sensory stimuli. Since we do not believe N-MNIST to encode discriminative features in time, we could then exploit such an artefact to do a rate-based classification, as we rightfully demonstrate in Section \ref{sec:popRate}. 

Why do we need a dataset that is discriminative in the time domain? Why not perform spike timing computations on static datasets (such as MNIST)? The spirit of neuromorphic engineering is not to just reproduce what deep learning has already done and make it more efficient, but to create a more brain-like method of computing – this should start from the dataset, continue at neural and learning levels, and finally produce brain-like functionality. As seen in the introduction of this paper, there is a lot of biological evidence that precise spike times play an important role in neural computations. The brain works on spatiotemporal patterns. SNNs use spikes as their units of computation. STDP uses difference between spike times as its measure for learning. To highlight the utility of these computational mechanisms, we need a dataset wherein features are encoded in individual spike times asynchronously.

Therefore, the performance of many neuromorphic algorithms on N-MNIST (and N-Caltech101) should be taken with a pinch of salt – this includes our algorithm. While we (and others) are able to successfully filter out noise induced by camera movements in N-MNIST, this in itself is not sufficient to assess the goodness of a neuromorphic algorithm. This brings us to the third and most important question – what constitutes a good neuromorphic dataset? The method of moving images or a vision sensor across static images in a Computer Vision dataset was one of the first attempts at creating a neuromorphic dataset. Clearly, in N-MNIST and N-Caltech101 at least, this does not seem to be sufficient at creating an appropriate dataset. If it is not sufficient, then what is? A useful candidate may be an audio dataset. Audio is inherently spatiotemporal, and summing up temporal events over time will result in huge loss of information. Audio dataset also does not have one single peak in amplitude that is representative of all patterns. Audio over a short duration also does not make sense. On the contrary, audio events are dynamic, and events that unfold over a period of time lead to a holistic representation of the information, as described in \cite{George:2008}. 

It may not be just audio that is inherently spatiotemporal, but \cite{Fiser:2002}, \cite{George:2008} argue vision is too. Although we are able to recognize a static image perfectly well, we are also able to generalize in a way that deep learning cannot – over different rotations, lighting conditions, sizes, and so on. This is possibly because we are exposed to a continuous stream of varying data (\cite{Simoncelli:2003}, \cite{Blake:2005}, \cite{Mazzoni:2011}, \cite{Faive:2014}, \cite{Keitel:2017}), and use time as a supervisor to understand and perform these generalizations (\cite{George:2008}). A visual dataset that embodies these principles may be suitable. For example, we may consider a dataset with retinal saccades over a 3 dimensional object with varying conditions – in this case, considering events over a short duration, or summing up events over time will not give anything useful. However, considering the changes along with the precise time information will lead to holistic representations not otherwise possible with static information. 

In conclusion, spikes occurring over time is not just an alternate mechanism for representing static information, such as using the intensity of a pixel as the rate for a Poisson spike train. Brains have evolved to use computing mechanisms that are inherently suitable to represent and process information from a dynamic world. A suitable benchmark should assess these computing mechanisms. In this paper, therefore, we address an important issue in neuromorphic computing by assessing N-MNIST and N-Caltech101 along the time domain, delineating why this is important. We also evaluate the requirements for a neuromorphic dataset. This in turn, highlights a need for further research into effective benchmarks and criteria for assessing the strengths of SNNs over earlier neural networks.  

\section*{Funding}

This research is supported by Programmatic grant no. A1687b0033 from the Singapore government’s Research, Innovation and Enterprise 2020 plan (Advanced Manufacturing and Engineering domain)

\section*{Conflicts of Interests Statement}

The authors declare that the research was conducted in the absence of any commercial or financial relationships that could be construed as a potential conflict of interest. 

\section*{Acknowledgements}

We would like to acknowledge our colleague Roshan Gopalakrishnan for working on N-Caltech101 collapsed images and improving the results. His contributions are a part of another paper (\cite{Gopalakrishnan:2018}) by our group.

\nolinenumbers

%This is where your bibliography is generated. Make sure that your .bib file is actually called library.bib
\bibliography{neuroBib}

%This defines the bibliographies style. Search online for a list of available styles.
\bibliographystyle{abbrv}

\end{document}